\documentclass[letterpaper, 10 pt, conference]{ieeeconf}

\IEEEoverridecommandlockouts                              

\usepackage{graphics} 
\usepackage{epsfig} 
\usepackage{mathptmx} 
\usepackage{times} 
\usepackage{amsmath} 
\usepackage{amssymb}  
\usepackage{graphicx}
\usepackage{tabu}
\usepackage{capt-of}
\usepackage{graphicx}
\usepackage{caption}
\usepackage{subcaption}
\usepackage{siunitx}
\usepackage{balance}
\usepackage[font=scriptsize,labelfont=bf]{caption}

\usepackage{tabularx}
\usepackage{float}
\usepackage{units}
\usepackage{url}
\usepackage{physics}
\usepackage{cite}

\title{\LARGE \bf
Pre-Clustering Point Clouds of Crop Fields Using Scalable Methods
}
\author{Henry J. Nelson and Nikolaos Papanikolopoulos\\
\footnotesize{$\{$nels8279, papan001$\}$@umn.edu} \\
\it \normalsize Department of Computer Science and Engineering, University of Minnesota \\
}

\begin{document}

\maketitle
\thispagestyle{empty}
\pagestyle{empty}
\begin{abstract}
In order to apply the recent successes of machine learning and automated plant phenotyping on a large scale using agricultural robotics, efficient and general algorithms must be designed to intelligently split crop fields into small, yet actionable, portions that can then be processed by more complex algorithms. In this paper, we notice a similarity between the current state-of-the-art for separating corn plants and a commonly used density-based clustering algorithm, Quickshift. Exploiting this similarity we propose a number of novel, application-specific algorithms with the goal of producing a general and scalable field segmentation algorithm. The novel algorithms proposed in this work are shown to produce quantitatively better results than the current state-of-the-art while being less sensitive to input parameters and maintaining the same algorithmic time complexity. When incorporated into field-scale phenotyping systems, the proposed algorithms should work as a drop-in replacement that can greatly improve the accuracy of results while ensuring that performance and scalability remain undiminished.
\end{abstract}

\section{Introduction}
While the demand for food increases and the need for environmentally conscious farming practices is clearer than ever before, precision agriculture and agricultural robotics promise the ability to reduce input costs, while simultaneously maximizing yield and reducing environmental impact. But to live up to these promises, phenotype data must be collected for individual plants on an extremely large scale \cite{htpp}.
The current solution, which is to estimate the phenotype of an entire field from a manually measured sample, is prone to sampling bias and does not scale. Requiring human intervention to produce data slows the rate at which data can be collected and makes every data point costly in terms of both time and money.
The remote sensing and agricultural robotics communities have been working to automate this process, but obtaining accurate measurements of individual plants in an unstructured environment is a multifaceted problem with many challenging intermediate steps. Currently proposed solutions utilize specialized sensors and computationally expensive algorithms that make these solutions unable to perform at the scale of entire fields. But instead of performing all operations on a large scale, it may be possible to collect data on a large scale using a robotic platform and then split the data into small, yet semantically meaningful, units that can be processed individually. This would allow data to be captured at a large scale and then broken into individual plants to be processed by a more computationally expensive phenotyping system, thus avoiding the need to feed large amounts of data into phenotyping systems while simultaneously simplifying the task they need to accomplish.
It is therefore necessary to find a way of splitting a field of plants into individual plants in a computationally inexpensive way.

\begin{figure}[t]
    \centering
    \includegraphics[width=0.48\textwidth]{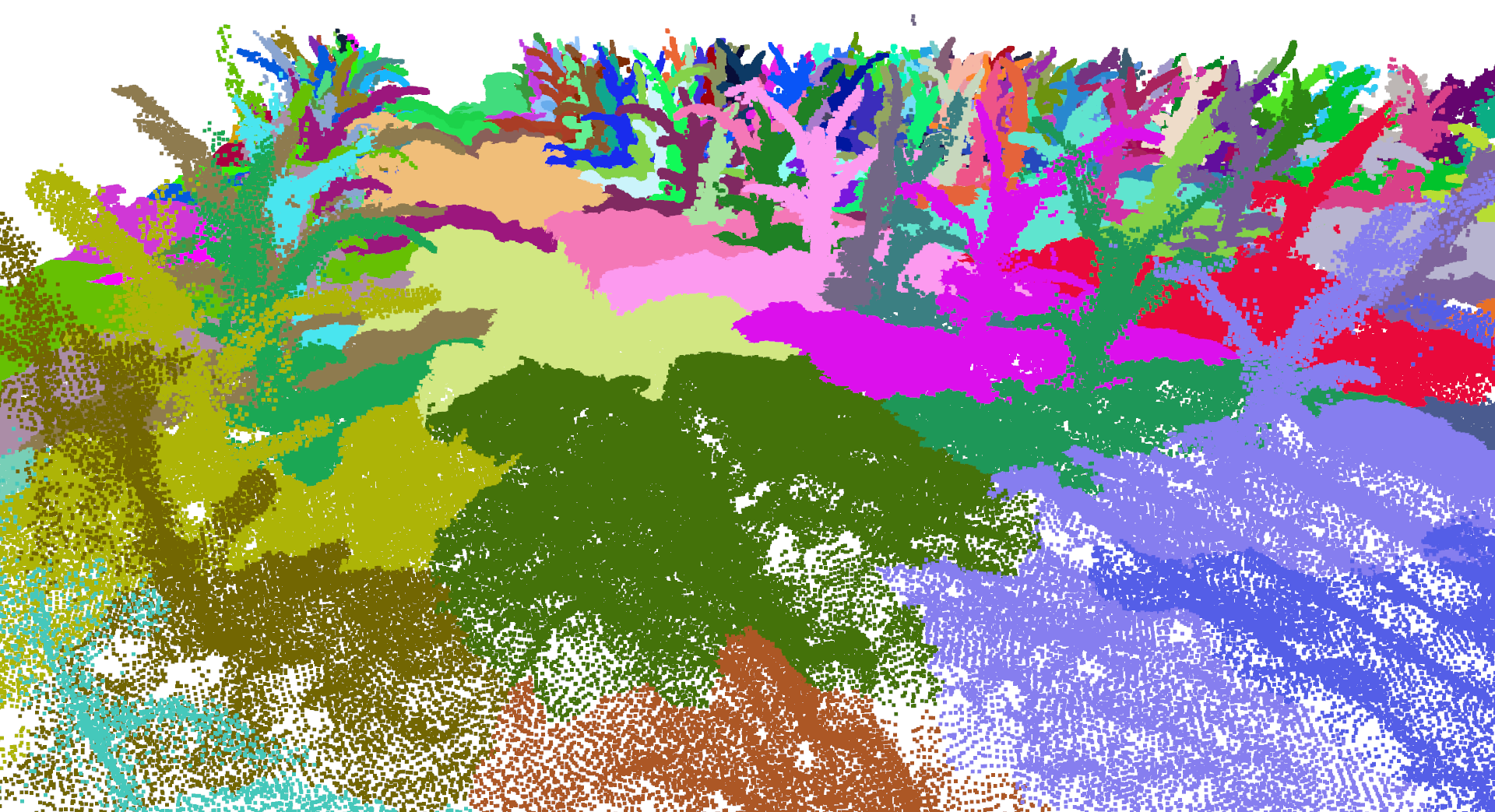}
    \caption{The result of Ground Density Quickshift++ when run on a point cloud of a corn field. Ground Density Quickshift++ is a novel algorithm designed to be general enough to apply in multiple applications and scalable enough to be used on entire fields while requiring no training data. A total of 167 clusters are found with only one case of a cluster containing more than one plant (shown in Fig. \ref{fig:double_plant}). There are 136 plants in the input point cloud meaning that some clusters, like those visible between rows in the center of this image, contain no plants.}
    \label{fig:teaser}
\end{figure}

In this paper, we present and evaluate a number of algorithms developed to roughly segment individual plants from one another over an entire field. The algorithms are all defined as slight modifications of a commonly used clustering technique so that their computational complexity is low and well defined.
The use of common data clustering techniques as an algorithmic basis constrains the new algorithms to be relatively data and species agnostic while maintaining a known behavior when scaling up to large input sizes.
These algorithms allow data collection and phenotype measurement to happen at different scales so the computationally expensive phenotyping algorithms can operate on small portions of the point cloud containing individual plants in order to maintain a lower runtime.

Because the plant separation problem is just a small part of what is to be a long and complex system for phenotype measurement, the required inputs and outputs of each algorithm are kept as general as possible. Each algorithm takes in a point cloud of a group of plants and produces a partition of the input points into sets \textit{roughly corresponding} to individual plants. It is important to note that since this step is meant as a pre-clustering step to produce clusters that will be further processed, the partitioning of plants does not have to be precise and a rough approximation is adequate as long as the algorithm scales. The input point clouds are required to be rotated such that the gravity vector of the world frame is pointing in the $-\Vec{z}$ direction (up is $+\Vec{z}$ direction), but the coordinate system is otherwise unconstrained. They do not require row, color, or radiometric information. Only the three spatial dimensions are necessary for each point.

The algorithms defined in this work are more general than any other method of separating plants in point clouds of crop fields. Their independence from coordinate system, training data, row information, and radiometric or color information make them the most widely applicable algorithms available for the task of plant segmentation. They do not require a specific sensor like lidar or an RGB camera and they do not require any training data like common machine learning approaches. One of them is even unit agnostic, avoiding the need to be scaled into known units as is commonly required for reconstructions created via structure-from-motion. The algorithms defined here are not the final solution to the problem they address, but they beat the current state-of-the-art by a large margin, are scalable enough to be used on large-scale data, general enough to be a starting point for most similar plant species, and simple enough to be adapted to specific problems.

\section{Related Work}

For segmenting a field into individual plants, specialized sensors like lidar have proven useful \cite{lidar, lidar_FRCNN, lidar_VCNN, lidar_cluster}, but lidar sensors remain prohibitively expensive when compared to more commonly available sensors like RGB cameras, which are already commonly used for crop surveying. In \cite{lidar}, radiometric data from the lidar sensor is directly utilized, meaning the algorithm is tied to a single sensing modality. In contrast, \cite{lidar_FRCNN} and \cite{lidar_VCNN} do not use radiometric information so they can presumably be implemented with different sensors, but they instead utilize deep learning methods that require large amounts of training data and are unable to operate at an adequate resolution for field scale point clouds due to the use of voxel representations. Of the works using lidar, the work of \cite{lidar_cluster} is the most general as they use lidar only to obtain the point cloud and then use a region growing clustering method that incorporates row and plant spacing parameters to separate plants. The resulting plant detection algorithm performs well on simulated data but the performance deteriorates on data from real crop fields, achieving an average detection rate of only 60\%.

While using lidar produces a more accurate and complete reconstruction of the environment, it has not been the only approach to crop reconstruction. Some groups have instead opted to use stereo vision or structure-from-motion as they do not require expensive sensors. But due to the increased noise and non-uniform sampling imposed by these reconstruction techniques, the approach to the problem is usually slightly different. The work of \cite{voxel_lab} only operates on data collected in laboratory conditions, \cite{manual} uses manually separated plant data, and \cite{just_leaf} skips the plant level altogether and just segments individual leaves. In \cite{jay2015field} they claim the ability to segment well separated plants from field data, but the method of separating plants from one another is not detailed enough to be reproducible. The workflow proposed in \cite{rain} uses structure-from-motion and then a skeletonization procedure before finally clustering the point cloud in order to achieve sparse detection and sampled segmentations of plants in a large scale point cloud. This work achieves state-of-the-art results with scalable performance but does so on a skeletonized representation of the point cloud meaning the resulting segmentation is on a more coarse scale than the original point cloud. The algorithm used is easily applicable to the original point cloud though, if the skeletonization is not run and the algorithm is performed on the entire point cloud rather than on a sampling, the complexity is still only $\mathcal{O}(n\log n)$ which matches the complexity of all the algorithms proposed in this paper and achieves reasonable results. In fact, when extending the RAIN algorithm defined in \cite{rain} to run without randomization, a similarity with the Quickshift algorithm \cite{quickshift} was noticed and became the basis of this paper.

Quickshift \cite{quickshift} is a density-based, mode finding, clustering algorithm. It is similar to Mean-Shift \cite{meanshift} in that points are clustered with those that are drawn to the same mode, but it is not iterative and only evaluates the density function $\mathcal{O}(n)$ times. This makes Quickshift very efficient as its complexity is dominated by spatial neighborhood queries which can be performed quickly with the help of spatial data structures like a $k$D-Tree. With this paper we sought to exploit the similarity between RAIN and Quickshift to define a more accurate and generalized plant clustering algorithm. In this way we switch from using algorithms made specifically for a single application to using a slightly modified version of a more general and well vetted clustering algorithm.

\section{Algorithms}\label{sec:algorithms}
Noticing how RAIN \cite{rain} takes advantage of point cloud topology and approximates common density based clustering methods, we define a few similar agglomerative clustering algorithms using Quickshift \cite{quickshift} as a template. The first of these is a non-randomized adaptation of the RAIN algorithm to be used for comparison while the others are slight modifications of Quickshift or Quickshift++ \cite{quickshift++}.

\begin{figure}[t]
    \centering
    \includegraphics[width=0.45\textwidth]{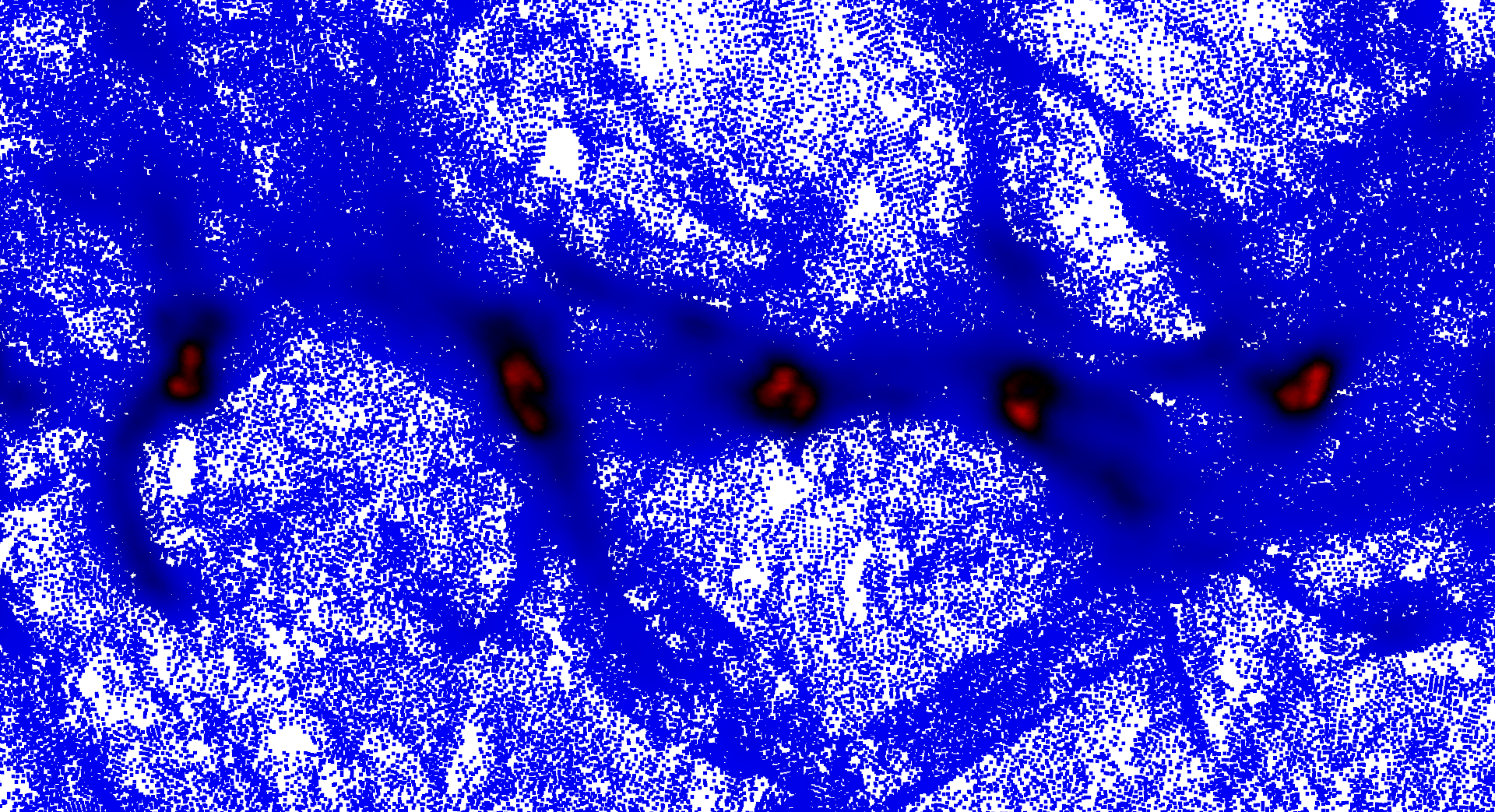}
    \caption{A visualization of a density function on a point cloud of plants projected into the ground plane. The red regions (areas of high density) are easily separable and identify individual plant stems. This is due to the vertical structure of the stem causing all points along the stem project to the same region of the ground plane.}
    \label{fig:2Dsensity}
\end{figure}

\subsection{Non-random RAIN}
The first algorithm we define is a non-randomized version of the RAIN algorithm from \cite{rain}. The purpose being that RAIN is the current state-of-the-art in this domain but it only partitions a sampling of the points and therefore does not directly match the formulation used in this work. So with this algorithm we remove the random sampling portion of the algorithm and run the procedure on all points in the point cloud. In this way it is directly comparable to the other algorithms in this work while maintaining its state-of-the-art performance. The algorithm takes as input a neighborhood distance $d$ and a point cloud $P=\{\Vec{p_1},\Vec{p_2},\hdots,\Vec{p_n}:\Vec{p_i}\in\mathbb{R}^3\}$. For each point $\Vec{p}\in P$, we assign $\Vec{p}$ to the cluster of the point in $N_d(\Vec{p})$ with the minimum $z$ coordinate where $N_d(\Vec{p}) = \{\Vec{n}\in P : \norm{\Vec{p}-\Vec{n}} < d\}$, the neighborhood of points within distance $d$ of the point $\Vec{p}$. Assuming this neighborhood can be computed with a spatial data structure like a $k$D-tree in approximately $\mathcal{O}(\log n)$ time, this means the entire algorithm operates in $\mathcal{O}(n\log n)$. This adaptation of the algorithm leaves it with only one parameter, $d$, which is a neighborhood distance threshold in 3D space.

\subsection{Z-Quickshift}
Since the RAIN algorithm clusters each point with its lowest ($\Vec{z}$ dimension) neighbor within some neighborhood, we can instead define an algorithm to cluster each point with its \textit{nearest} lower neighbor. This modification preserves the topology following analogy used in \cite{rain} of a raindrop running down a plant while formulating the algorithm in a way that fits the mold of Quickshift \cite{quickshift}. Similar to non-randomized RAIN, this algorithm maintains the single distance parameter $d$ governing neighborhood searches. This algorithm can be implemented as a simple modification of the density function of Quickshift to $density(\Vec{p}) = -\Vec{p}\cdot\Vec{z}$. The modified density function is extremely simple to compute and does not require neighborhood information, but it does not change the computational complexity of Quickshift which is dominated by neighborhood finding. The modified Quickshift is dubbed \textit{Z-Quickshift} in this work and also has a time complexity of $\mathcal{O}(n\log n)$.

\subsection{Ground Density Quickshift}
Instead of utilizing the vertical structure of the plant, it was noted that viewing the structure of the plant in the horizontal plane gives a very clear picture of the separation between plants and takes advantage of the horizontal density common to many vertically structured plants (corn, trees, hops) to cluster points that are close in the projected space. This structure can be seen in Fig. \ref{fig:2Dsensity}. This is conducive to clustering points that are close in the projected space but may not be close in 3D, such as two points on opposite ends of the plant stem or trunk. To take advantage of the observable structure in the ground plane we can project all points into the ground plane where we can then perform standard Quickshift in 2D. For this algorithm we utilize the density function defined in \cite{quickshift++} as it can be performed with the neighborhood queries which must be computed for Quickshift anyway, allowing a runtime of $\mathcal{O}(n\log n)$. The density function has a single integer parameter $k$ determining the kernel size. This gives \textit{GD Quickshift} a total of two parameters, $d$ and $k$.

\subsection{Ground Density Quickshift++}
Projecting the data into the ground plane before clustering easily separates vertical structures but will also place leaves in the same cluster if they overlap in the 2D space even if they are far apart in 3D. To combat this we could instead use the densities determined in 2D and cluster the points in 3D, but this creates many small clusters along the length of the stem rather than keeping them all in the same cluster. As a workaround, we can instead use Quickshift++ \cite{quickshift++} and cluster the densest regions of nearby points in the 2D density space as an initialization. We can then cluster the remaining points in 3D to ensure points are only in the same cluster if they are close in 3D or were in an extremely dense region in the 2D projection (probably a stem). Due to the initialization, there is no need for a distance parameter making this algorithm completely agnostic to scale. Instead, Quickshift++ uses a parameter $\beta\in[0, 1]$ to identify the densest regions of points for the initialization. Along with the $k$ parameter used for the density function, \textit{GD Quickshift++} has a total of two parameters. The initialization of clusters runs in $\mathcal{O}(n\log n)$ so the final algorithm is also $\mathcal{O}(n\log n)$.

\section{Data}\label{sec:data}
To evaluate the algorithms from Section \ref{sec:algorithms}, both synthetic data and real-world sensor data are used. The corn field data and implementations of the algorithms have been made publicly available\footnote{\url{https://github.com/hennels/CropPreClustering}} to facilitate the replication of results and continued development by the community. The non-corn data used in this work are already publicly available \cite{Fuji}.

Synthetic data was created by taking the individual plant reconstructions provided in \cite{corn50} and organizing them into a large scale field of plants. This allows us to have a known ground truth partitioning of the point cloud with which to quantitatively compare against. Two large scale point clouds are synthesized, one containing 100 plants and the other containing 400 plants. The ground truth partitions are encoded in the point clouds using color information.

\begin{figure}[t]
    \centering
    \includegraphics[width=0.48\textwidth]{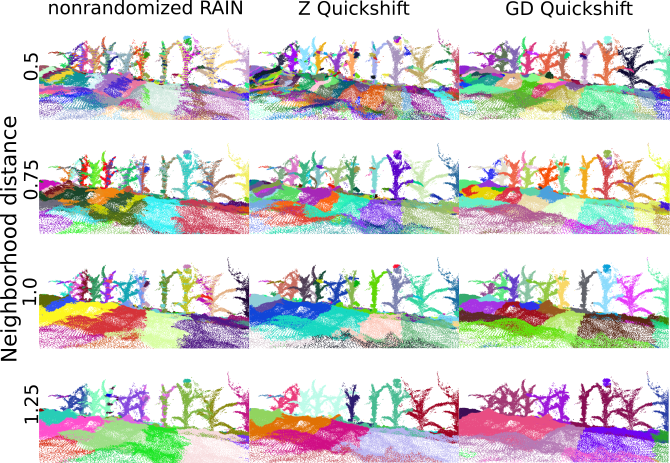}
    \caption{Qualitative results for a portion of the St. Paul V3 data. Each column is a different algorithm and each row shows a different value of the neighborhood distance threshold $d$. The equivalent image for Quickshift++ is shown in Fig. \ref{fig:stpaulv3_pp}.}
    \label{fig:stpaulv3}
\end{figure}

A small dataset of reconstructions from production corn fields was also compiled. A UAV with an attached RGB camera flew at a low altitude over corn fields collecting video. Afterwards, the video was used to build a 3D reconstruction of the corn plants using Pix4D to perform structure-from-motion. The dense central region of each reconstruction was then extracted and transformed such that the plant stems were parallel with the $\Vec{z}$-axis. The data was collected in three different fields over two different growing seasons with corn plants spanning stages from V3 to V6. Qualitative results from only one point cloud are shown in this paper (in Figures \ref{fig:teaser}, \ref{fig:stpaulv3}, and \ref{fig:stpaulv3_pp}) but the reader is encouraged to test the algorithms on the other provided point clouds to see additional results.

\section{Evaluation}\label{sec:eval}
\begin{table}[b]
\centering
\begin{tabular}{|l|l|l|l|}
\hline
\multicolumn{1}{|c|}{Algorithm} & \multicolumn{1}{c|}{$d$} & \multicolumn{1}{c|}{$k$} & $\beta$ \\ \hline
\text{Non-random RAIN} & \checkmark &  &  \\ \hline
\text{Z Quickshift} & \checkmark &  &  \\ \hline
\text{GD Quickshift} & \checkmark & \checkmark &  \\ \hline
\text{GD Quickshift++} &  & \checkmark & \checkmark \\ \hline
\end{tabular}
\caption{The different parameters required by each algorithm. $d$ is a neighborhood distance threshold and is thus dependant on the scale of the reconstruction. $k$ is a kernel size parameter for a density function and $\beta$ is a multiplicative factor for the density function so both are dependant on the density of points in the point cloud.}
\label{tab:parameters}
\end{table}

In order to evaluate the algorithms proposed in this paper we use both quantitative and qualitative methods. For the synthetic data with ground truth partitions we are able to directly compare the results of the algorithms with the ground truth after finding a bipartite correspondence between predicted and ground truth clusters. To compare partitions the intersection over union (IoU) is used and a bipartite matching between predicted and ground truth clusters is computed such that the sum of IoUs is maximized. This makes sure each estimated cluster is only compared against a single ground truth cluster. For the real-world data where ground truth is not available, we show qualitative results for each algorithm using a variety of parameters. To evaluate, we run each algorithm on the same point cloud, with the same parameters, and view the differences. Since each algorithm requires different parameters (see Table \ref{tab:parameters}), we notice that many of the algorithms are most sensitive to the neighborhood distance threshold $d$, so this is the parameter we vary. For $k$, a few values were tried and one with reasonable performance on all point clouds ($k=1200$) is shown. The GD Quickshift++ algorithm was found to give good results for any value of $\beta\in[0.1, 0.4]$ so a value of $\beta=0.3$ is used for all work shown in this paper. For the best performing algorithm we are able to compare manually verifiable metrics on real-world data such as the ground truth number of plants in the point cloud vs the total number of clusters and the number of clusters containing more than one plant because the errors are few enough to be manually counted.

Since these are pre-clustering algorithms, it is important to note that the necessary performance is dependant on what the subsequent processing algorithms can handle. Ideally, a plant pre-clustering algorithm should be able to separate plants and keep all plants whole in the process, but these goals are often in opposition to one another and it is difficult to accomplish both well. So depending on the subsequent processing algorithm, an over or under-fragmented segmentation may be preferable and the parameters could be adjusted to produce such a partition.

\subsection{Quantitative Evaluation}

\begin{table}[t]
\centering
\begin{tabular}{|c|c|c|c|}
\hline
Algorithm       & Clusters     & mean IoU        & median IoU      \\ \hline
Ground Truth    & 400          & -               & -               \\ \hline
Non-random RAIN & 474          & 62.1\%          & 68.4\%          \\ \hline
Z-Quickshift    & 419          & 62.9\%          & 72.7\%          \\ \hline
GD Quickshift   & 397          & 81.0\%          & 81.7\%          \\ \hline
GD Quickshift++ & \textbf{401} & \textbf{84.3\%} & \textbf{85.1\%} \\ \hline
\end{tabular}
\caption{Quantitative results on \texttt{10to29\_1M.ply} for each algorithm. A single IoU value is computed between each predicted and ground truth cluster that are matched by the bipartite matching. The mean and median values of these matched pairs are reported..}
\label{tab:iou1}
\end{table}

\begin{table}[b]
\centering
\begin{tabular}{|c|c|c|c|}
\hline
Algorithm       & Clusters     & mean IoU        & median IoU      \\ \hline
Ground Truth    & 100          & -               & -               \\ \hline
Non-random RAIN & 102          & 54.9\%          & 65.8\%          \\ \hline
Z-Quickshift    & 102          & 64.9\%          & 81.3\%          \\ \hline
GD Quickshift   & 116          & 81.9\%          & 82.7\%          \\ \hline
GD Quickshift++ & \textbf{100} & \textbf{85.7\%} & \textbf{87.1\%} \\ \hline
\end{tabular}
\caption{Quantitative results on \texttt{0to9\_1M.ply} for each algorithm. A single IoU value is computed between each predicted and ground truth cluster that are matched by the bipartite matching. The mean and median values of these matched pairs are reported.}
\label{tab:iou2}
\end{table}

For quantitative evaluation on the synthetic data, each algorithm was run with a range of parameters. The results in Tables \ref{tab:iou1} and \ref{tab:iou2} show the best results for each algorithm that produced a number of clusters within 20\% of the ground truth value on the two synthetic point clouds. Due to the bipartite matching between estimated and ground truth clusters, the IoU values can be significantly biased if the number of estimated clusters differs significantly from the number of ground truth clusters which is the reason for the 20\% bound. In both Tables \ref{tab:iou1} and \ref{tab:iou2} Ground Density Quickshift++ outperforms the other algorithms on all metrics and thus this is the algorithm we also evaluate quantitatively on real-world data.
After applying Ground Density Quickshift++ to each of our real-world point clouds the number of plants and the number of clusters containing more than one plant were manually counted in each. The number of clusters with more than one plant is a metric measuring the cases where the algorithm was unable to separate plants. All of these occurrences were clusters with only two plants, most of which were double planting events as shown in Fig. \ref{fig:double_plant}. Double planting events are events where two seeds are planted in the same location because of a malfunction in the planting machinery. The number of clusters found minus the total number of plants in the point cloud is a measure of how many extraneous clusters the algorithm produces. Ideally, the number of multi-plant clusters and the number of extra clusters should both be zero. While we considered any cluster not containing a plant of the cultivated crop extraneous, it was noted that many of these clusters were on weeds in the field, so they may be desirable for applications such as weed monitoring or planning herbicide application. The results are shown in Table \ref{tab:quant}. From these values we can see that the total number of clusters produced by Ground Density Quickshift++ is only slightly higher than the total number of plants and the number of multi-plant clusters is relatively small for all three point clouds.

\begin{table}[t]
\centering
\begin{tabular}{|c|c|c|c|}
\hline
Dataset & Total Plants & Total clusters & Multi-plant Clusters \\ \hline
St. Paul V3 & 136 & 144 & 4 \\ \hline
St. Pual V6 & 103 & 107 & 6 \\ \hline
Dji V4 & 90 & 92 & 3 \\ \hline
\end{tabular}
\caption{Quantitative results for GD Quickshift++ running on three different point clouds. Note that the parameters of GD Quickshift++ were not tuned for any one of these point clouds, the values of $k=1200$ and $\beta=0.3$ were chosen as parameters that work reasonably well for all of them. Improvement for each would be expected with parameter tuning.}
\label{tab:quant}
\end{table}

\subsection{Qualitative Evaluation}
\begin{figure}[t]
    \centering
    \includegraphics[width=0.46\textwidth]{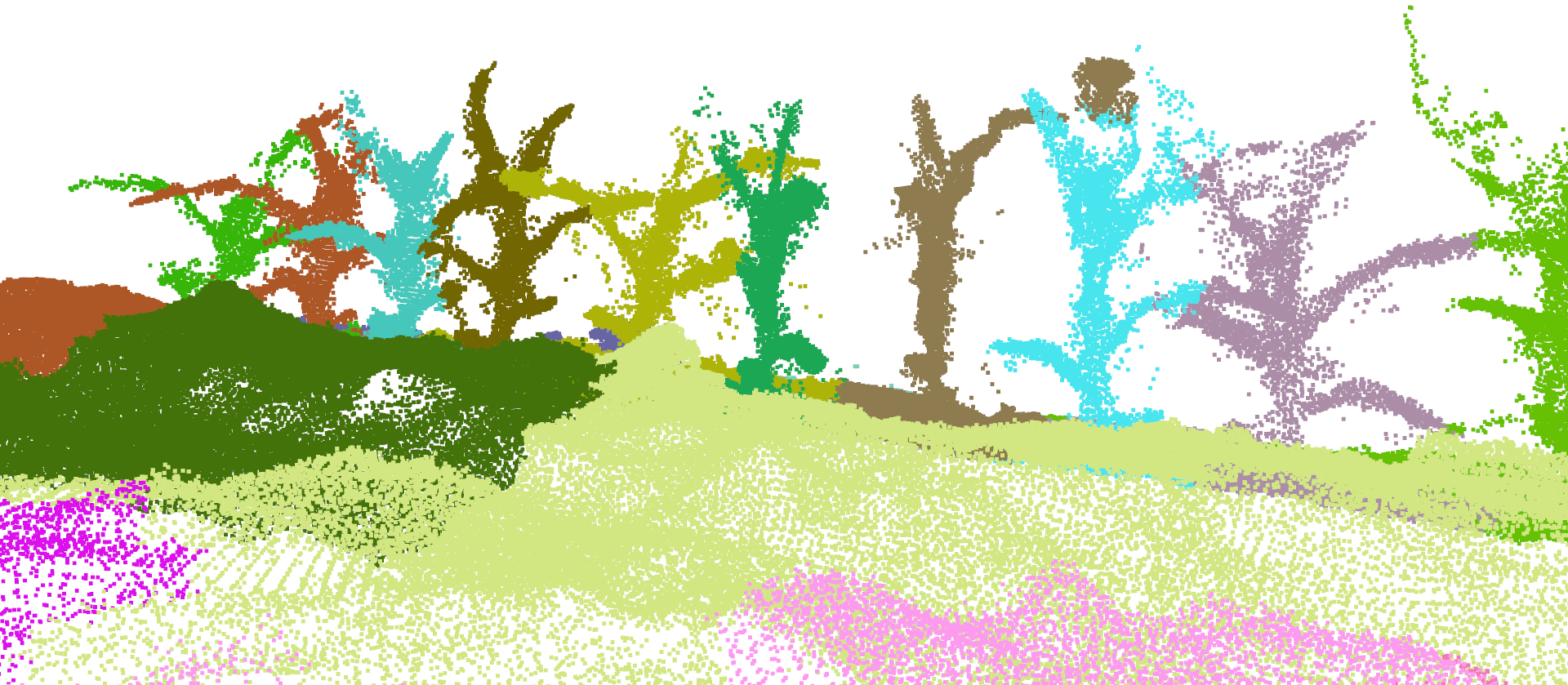}
    \caption{Qualitative results for Quickshift++ on a portion of the St. Paul V3 data. It is directly comparable to the results in Fig. \ref{fig:stpaulv3} but does not use a neighborhood distance threshold. See Fig. \ref{fig:teaser} for a view of the whole field.}
    \label{fig:stpaulv3_pp}
\end{figure}

In Figures \ref{fig:stpaulv3} and \ref{fig:stpaulv3_pp} qualitative results on the St. Paul V3 dataset are shown. From these images it is clear that non-randomized RAIN (left column of Fig. \ref{fig:stpaulv3}) is unable to completely separate plants with any of the sampled parameters and often produces more than one cluster on each plant. The behavior changes quickly from over-fragmentation to under-fragmentation as the neighborhood distance threshold increases. Z Quickshift (center column of Fig. \ref{fig:stpaulv3}) gives slightly cleaner segmentations but maintains the fragmentation sensitivity to the distance parameter. Ground Density Quickshift (right column of Fig. \ref{fig:stpaulv3}) gives comparatively much cleaner segmentations and seems less sensitive to the neighborhood distance threshold as the over and under-fragmentation are well balanced for a majority of the $d$ input values. The Ground Density Quickshift++ result, which is shown in Fig. \ref{fig:stpaulv3_pp} as it does not utilize a neighborhood distance threshold, is the cleanest of any of the algorithms tested and is very insensitive to input parameters. Each plant in the Ground Density Quickshift++ result is part of just a single cluster while the ground points are placed in relatively few clusters. Performance of Ground Density Quickshift++ on a larger portion of the same dataset can be seen in Fig. \ref{fig:teaser}.

\begin{figure}[b]
    \centering
    \includegraphics[width=0.48\textwidth]{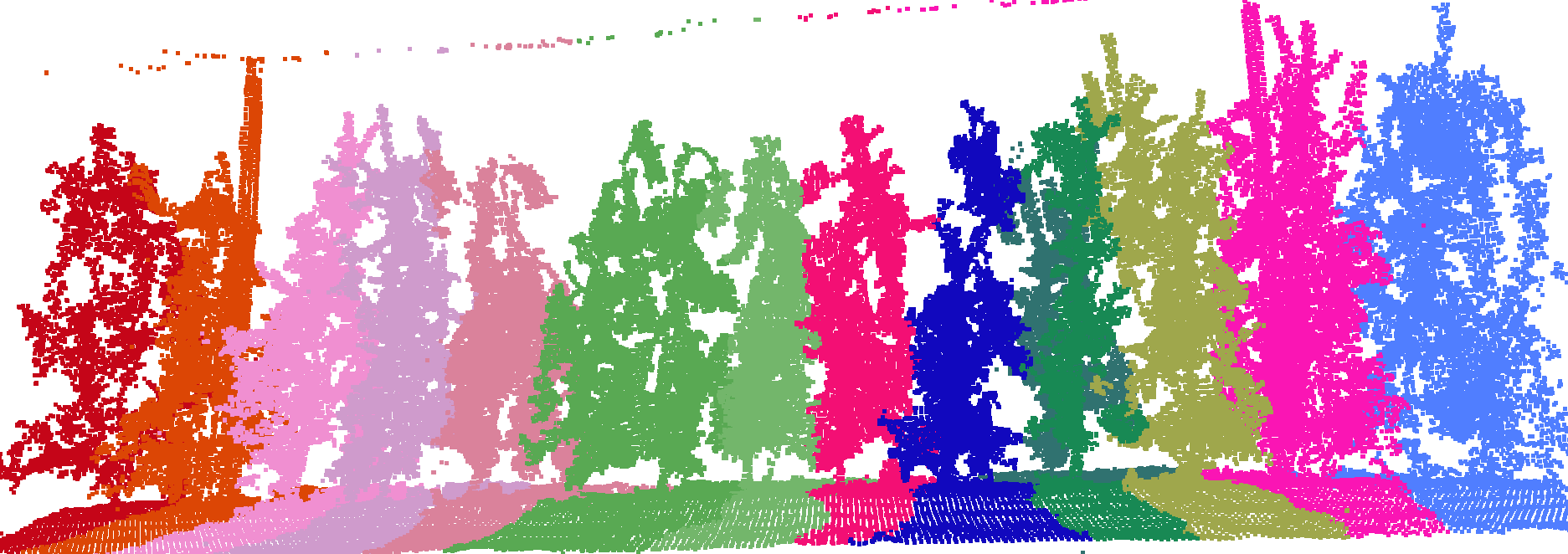}
    \caption{The result of running Quickshift++ on structure from motion data of Fuji apple trees from \cite{Fuji}. Clusters mostly represent individual trees. Minor errors in segmentation balance out so that the total number of trees detected is correct.}
    \label{fig:apple_trees}
\end{figure}

While this work focuses on corn segmentation, the algorithm shows promise for a variety of crops. As mentioned previously, during experiments on data from corn fields a number of weeds were found to also be well segmented by the algorithm. To investigate whether the algorithms are effective on other crop types, Ground Density Quickshift++ was run on a a publicly accessible point cloud of Fuji apple trees \cite{Fuji}. The result is visible in Fig. \ref{fig:apple_trees}. The algorithm was able to produce a reasonable segmentation of the apple trees in the point cloud, making only a few small errors. Two of the apple trees were unable to be separated and one side of a tree that was particularly dense with apples was counted as a separate tree, making the total number of clusters equal to the number of trees captured in the reconstruction. While the apple tree dataset is a small example, it and the weeds that were segmented from the corn datasets do suggest that Ground Density Quickshift++ is extensible to some other crop domains with little to no modification.

\section{Discussion}
\begin{figure}[t]
    \centering
    \includegraphics[width=0.2\textwidth]{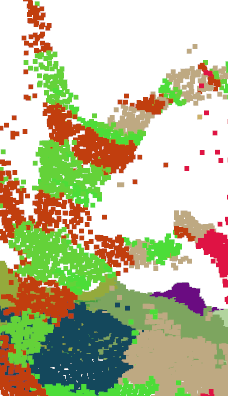}
    \includegraphics[width=0.2\textwidth]{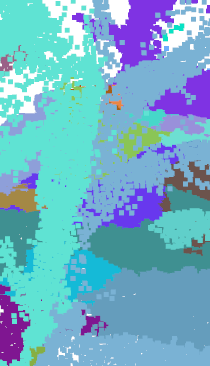}
    \caption{Common failures of (\textbf{Left}) non-random RAIN and (\textbf{Right}) Z-Quickshift. Since RAIN is always looking for the lowest point, it can produce layered clusters on the same plant. While Z-Quickshift cannot produce a horizontally layered structure, it can produce vertical layers for each side of the plant.}
    \label{fig:periodic_split}
\end{figure}

Although all the algorithms presented in this work have a similar structure, they have very different behaviors as is seen in Section \ref{sec:eval}. The non-random RAIN algorithm does not group nearby points well because it's goal groups points with their lowest neighbor within some distance. This does not ensure that points cluster with their immediate neighbors and often results in a periodic clustering pattern as shown on the left side of Fig. \ref{fig:periodic_split}. Similarly, Z-Quickshift can also produce layered structures because it may not cluster points that are close together if there are different local minima at the bottom of the plant (Fig. \ref{fig:periodic_split} right). Both of these algorithms show the need for points to be clustered with neighbors as long as they are part of the same structure. This motivates the use of the density function used in the Ground Density algorithms. Both of the ground density function based approaches work much better in general and produce extremely clean clusters that mostly represent individual plants.

\begin{figure}[b]
    \centering
    \includegraphics[width=0.45\textwidth]{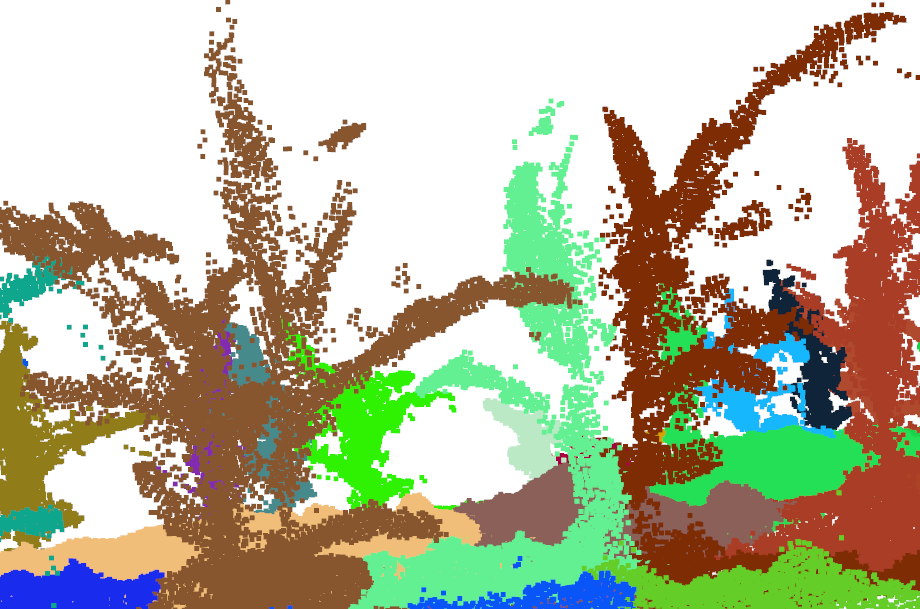}
    \caption{Two examples of double planting events. Plants from two different seeds that are planted in the same place. The example on the left shows an instance of close plants that the algorithm was unable to separate while the instance on the right is correctly separated.}
    \label{fig:double_plant}
\end{figure}

The main difference between Ground Density Quickshift and Ground Density Quickshift++ is the presence or absence of the neighborhood distance threshold. The neighborhood parameter governs how close two points need to be to be clustered together. This means it will have to be less than the distance between the nearest two plants if they are to be separated, but this is difficult in some situations like the double planting events shown in Fig. \ref{fig:double_plant}. The absence of the threshold in Ground Density Quickshift++ lets the threshold differ across the point cloud. Plants are not evenly distributed in a field, so typically the threshold would need to be lower in the row of plants and larger in the gap between rows to minimize the number of extraneous clusters. Requiring a constant threshold ensures that either plants will not be separated or there will be a large number of extraneous clusters. One of these alternatives may be permissible because they are able to be handled by subsequent algorithms but this deficiency adds algorithmic complexity later in the processing pipeline. Ground Density Quickshift++ does not require a constant threshold so it is better able to adapt to these different plant densities. This explains its good performance on the areas of different density seen in Figs. \ref{fig:stpaulv3_pp} and \ref{fig:teaser} where it has many clusters close together in the row of plants but very few in the gap between plant rows.

Without the explicit neighborhood distance threshold, Ground Density Quickshift++ is independent of the scale of the point cloud meaning a point cloud produced via structure-from-motion does not need to be scaled to world units before processing. The other three algorithms all require the neighborhood distance threshold and thus the choice of $d$ is dependant on the units of the point cloud. So in order to use consistent parameters, point clouds would need to be scaled to world units or produced using a method with known scale like lidar. Even though Ground Density Quickshift++ is independent of scale, it is dependent on the sampling density of the point cloud due to its density function. If scale or point cloud density is easier to control, this difference may inform which algorithm may be of use in a given application.

\section{Conclusion}
In this paper we proposed and evaluated a range of scalable clustering algorithms designed to give good results, even at the scale of entire fields. The best of these, Ground Density Quickshift and Ground Density Quickshift++, both perform better than the current state-of-the-art on scalable individual plant segmentation. While the performance differs slightly, all algorithms in this work have a time complexity of approximately $\mathcal{O}(n\log n)$. We also provided a synthetic dataset with ground truth segmentations and a real-world dataset on which to test the proposed algorithms and to foster further development in this domain. 

This work enables the development of a high throughput plant phenotyping system for agricultural robotics by providing algorithms to act as a first step to break a reconstruction into manageable parts representing individual plants; a task that any robotic system will have to accomplish in order to apply techniques from precision agriculture. This allows data capture to occur at field scale without requiring the subsequent phenotyping algorithms to operate at the same large scale, allowing them to be more complex and specialized. More specialized algorithms will need to be developed to operate on the resulting clusters, but many machine learning techniques have already shown promising results in such niche contexts. The main contribution of this work is not only that the proposed algorithms beat the current state-of-the-art method with the same time complexity, but also the generality and thus applicability of the proposed algorithms. As long as plants have a single vertical structure (like corn, fruit trees, or hops), one of the Ground Density algorithms should be able to find and segment them.

\section{Acknowledgements}
We would like to thank all members of the Center for Distributed Robotics at the University of Minnesota for help with data collection and fruitful discussion. A special thanks goes to Dimitris Zermas and Sentera, Inc. for processing imagery. USDA/NIFA has  supported this work through the grant 2020-67021-30755. This work was partially supported by the National Science Foundation through grants \#CNS-1439728, \#CNS-1544887, and \#CNS-1939033.

\typeout{}
\bibliographystyle{IEEEtran}
\bibliography{bib.bib}

\balance

\end{document}